\documentclass{article} 
\usepackage{iclr2026_conference,times}

\usepackage{amsmath,amsfonts,bm}

\def\eqref#1{equation~\ref{#1}}

\def\1{\bm{1}}

\DeclareMathAlphabet{\mathsfit}{\encodingdefault}{\sfdefault}{m}{sl}
\SetMathAlphabet{\mathsfit}{bold}{\encodingdefault}{\sfdefault}{bx}{n}

\usepackage{booktabs}   
\usepackage[table]{xcolor} 
\usepackage{amsmath}    
\usepackage{multirow}
\usepackage{wrapfig}
\definecolor{baselinegray}{gray}{0.95}
\definecolor{improvementblue}{cmyk}{0.1, 0, 0, 0}
\definecolor{textgray}{gray}{0.55} 
\definecolor{deltablue}{cmyk}{1, 0, 0, 0} 
\definecolor{classcolor}{HTML}{FCE5CD}
\definecolor{vqacolor}{HTML}{D9E1F2}
\definecolor{retcolor}{HTML}{E2EFDA}
\definecolor{grdcolor}{HTML}{F2DCDB}
\definecolor{finalcolor}{HTML}{DEEBF6}
\definecolor{oodcolor}{HTML}{FFF2CC}
\newcommand{\deltares}[1]{\textcolor{deltablue}{(#1)}}
\usepackage{hyperref}
\usepackage{url}
\usepackage{graphicx}

\usepackage{pifont}

\definecolor{mygreen}{rgb}{0.1, 0.6, 0.1} 
\definecolor{myred}{rgb}{0.8, 0.2, 0.2}   

\title{Explore More, Learn Better: Parallel MLLM Embeddings under Mutual Information Minimization}

\iclrfinalcopy

\author{Zhicheng Wang\textsuperscript{2,3}, \, Chen Ju\textsuperscript{1,2}\thanks{Corresponding Author and Project Leader}, \, Xu Chen\textsuperscript{2}, \, Shuai Xiao\textsuperscript{2}\thanks{Corresponding Author}, \, \textbf{Jinsong Lan\textsuperscript{2}}, \, \textbf{Xiaoyong Zhu\textsuperscript{2}}, \\
\textbf{Ying Chen\textsuperscript{2}}, \,
\textbf{Zhiguo Cao\textsuperscript{3}} \\
\textsuperscript{1} Zhejiang University, 
\textsuperscript{2} Alibaba Group, 
\textsuperscript{3} Huazhong University of Science and Technology \\ 
\texttt{\{zhicheng\_wang,zgcao\}@hust.edu.cn} \quad   \texttt{cju.void@gmail.com} 
}

\begin{document}

\maketitle
\vspace{-25pt}
\begin{center}
\url{https://github.com/Xu3XiWang/PDF-VLM2Vec}
\end{center}
\begin{abstract}
Embedding models are a cornerstone of modern AI. Driven by Multimodal Large Language Models (MLLMs), they have made great progress in architecture and data curation, while the holistic paradigm is still limited to SSC, {\em i.e.}, single input, singular embedding, contrastive supervision, which collapses rich, multifaceted inputs into monolithic embeddings and fails to fully exploit MLLM capabilities. In this paper, we tailor one \textbf{P}arallel \textbf{D}ecoupling \textbf{F}ramework (PDF) for multimodal embedding learning, by utilizing the proprietary steerability of MLLMs, {\em i.e.}, their ability to flexibly generate quite differentiated response under explicit instructions.
Concretely, PDF conditions a shared MLLM backbone on distinct, learnable prefixes to roll out multiple parallel paths for one input, then relies on these paths to obtain parallel embeddings. To promote full parallel diversity, we employ Mutual Information Minimization (MIM) as an explicit constraint, coupled with per-path contrastive supervision to maintain semantic alignment. 
Such dual-objectives force PDF to yield robust semantic coverage and a generalizable embedding space. Ultimately, the remarkable embedding space are accessible at inference via one single forward pass, incurring negligible computational overhead.
We instantiate PDF on multiple MLLM backbones and prove its effectiveness on MMEB benchmark. Significant gains are consistently achieved across various resolutions and model sizes, {\em e.g.}, boosting the VLM2Vec-LLaVA-1.6-LR model by a remarkable +8.9\% (7B), while the VLM2Vec-Qwen2VL models by +4.2\% (2B) and +3.1\% (7B). In terms of efficiency, our 2B model surpasses its baseline by +2.6\% using only half the computational budget. 
\end{abstract}

\section{Introduction}
\begin{figure}[tp]
\centering
\vspace{-0.5cm}
\includegraphics[width=\linewidth]{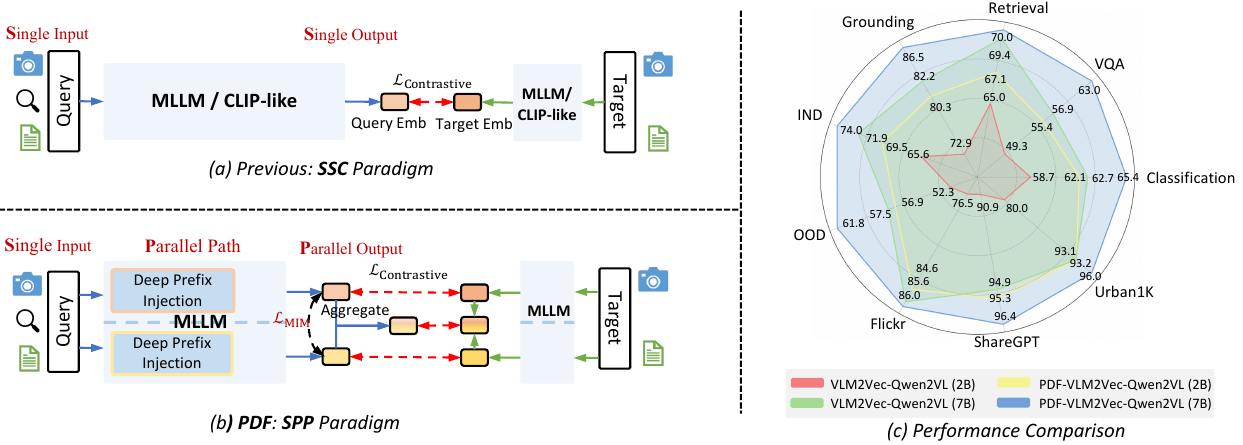}
\vspace{-0.7cm}
\caption{\textbf{Comparisons of Framework \& Performance.} (a) Previous: the ubiquitous SSC framework, {\em i.e.}, single input, singular embedding, contrastive supervision. 
(b) Parallel Decoupling Framework (PDF): single input rolls out as parallel inputs, then generates multiple embeddings explicitly de-correlated by Mutual Information Minimization (MIM). 
(c) On the VLM2Vec-Qwen2VL backbone, our PDF consistently delivers significant gains across diverse tasks and model scales.}
\label{fig:comparasion}
\vspace{-15pt}
\end{figure}

Embedding models, which encode complex inputs like text and images into dense vectors, are a cornerstone of modern AI, powering applications like semantic similarity~\cite{chechik2010large,agirre2012sem,marelli2014semeval,chen2025learningmultibranchcooperationenhanced}, information retrieval~\cite{lin2014microsoft,mitra2017learning,karpukhin2020dense,zheng2021contrastive} and Retrieval-Augmented Generation~\cite{lewis2020retrieval,izacard2020leveraging,guu2020retrieval,jin2025search}. To advance embedding models, previous efforts largely follow two main paths. On the data side, many studies~\cite{zhou2024megapairs,chen2025mme5,gu2025breaking,lan2025llave} explored labor-intensive hard-sample mining. On the architectural side, early CLIP-like models, {\em e.g.}, UniIR on the M-BEIR benchmark~\cite{lan2025llave} have evolved into recent MLLM-based models like VLM2Vec on the MMEB benchmark~\cite{jiang2024vlm2vec}, as Multimodal Large Language Models show impressive gains over small-scale counterparts. However, from the holistic perspective, most methods still converge on one ubiquitous \textbf{SSC} paradigm shown in Fig.~\ref{fig:comparasion}\,(a): mapping a \textit{\textbf{S}ingle input to a \textbf{S}ingular embedding, and learning via a \textbf{C}ontrastive supervision}. Such an identical paradigm appears increasingly outdated, mainly for two reasons. \textit{Architectural Mismatch}: originating from simpler CLIP-like dual-encoders, SSC fails to leverage the advanced capabilities of MLLMs. \textit{Information Bottleneck}: by collapsing a multifaceted input into a singular point in the embedding space, SSC incurs severe information loss, resulting in limited semantic coverage and reduced robustness.

This status quo drives us to tailor a proprietary embedding pipeline for MLLMs, enabling the learned embeddings to fully encompass semantic richness of input. Here, we are motivated by \textbf{\textit{the distinctive steerability}} of MLLMs~\cite{liu2023visual,liu2024improved,wang2024qwen2,qwen2025qwen25technicalreport} compared to CLIP-like dual-encoders, {\em i.e.}, MLLMs are highly flexible to prompts, tend to generate sufficiently differentiated response under explicit instructions. Prior work~\cite{jiang2024vlm2vec} mainly leverages such steerability for cross-task adaptation with the notable success. In contrast, we creatively utilize it for fundamental embedding learning. Concretely, the model is conditioned on distinct, learnable prefixes to roll out a single input into multiple parallel prompts, which are subsequently processed by MLLMs to yield multiple parallel embeddings. This constitutes our novel \textbf{SPP} paradigm: \textit{\textbf{S}ingle input, \textbf{P}arallel paths, \textbf{P}arallel outputs}. For the same input, prefixes are initialized differently, leading parallel rollouts to follow divergent convergences, thus achieving robust semantic coverage.

To better realize our \textbf{SPP}, naively relying on randomly initialized prefixes to \textit{implicitly} guide parallel paths is insufficient and suboptimal. Although prefixes differ at initialization, the shared MLLM backbone and mere one strong objective (contrastive loss) create a clear tendency for the model to gradually disregard differentiation during convergence, causing all parallel paths to collapse into somehow similar, {\em i.e.}, redundant embeddings. To counteract this, we argue that an \textit{explicit} constraint on parallel differentiation is necessary, {\em i.e.}, actively measure then penalize statistical dependencies between parallel paths. This force embeddings to become diverse, maximizing semantic coverage to minimize information loss, moving beyond the singular-embedding limitation of \textbf{SSC}.

As illustrated in Fig.~\ref{fig:comparasion}\,(b), we hence propose one novel \textbf{Parallel Decoupling Framework (PDF)}, with threefold designs. First, to instantiate the parallel paths, we employ one \textit{\textbf{deep prefix injection}} mechanism: for each path, one unique set of learnable parameters is injected into every transformer layer, directly modulating the self-attention computation. Second, to fulfill the need for an explicit diversity constraint, we employ \textit{\textbf{Mutual Information Minimization}} (\textbf{MIM})~\cite{kinney2014equitability}. As the true data distribution is unknown, direct MI computation is intractable. We therefore minimize one tractable variational upper bound, implemented using the vCLUB estimator~\cite{cheng2020club}. This establishes a two-stage optimization game within each training step: a parametric MI estimator first learns to detect the dependency between parallel embeddings, after which the main MLLM is trained to produce more independent embeddings that ``fool" this fixed estimator. Third, to ensure this diversity does not degrade the embedding quality, a standard contrastive loss is applied to each path to anchor it to semantics of the input. This dual-objective training acts as a powerful regularizer on the shared model backbone~\cite{zhang2018deep}. Empirically, we find that a single forward pass at inference is sufficient to unlock the model's enhanced embedding space, {\em i.e.}, remaining efficient with negligible additional computational overhead.

To evaluate the effectiveness and generality, we instantiate PDF on top of the VLM2Vec paradigm, using both LLaVA-1.6~\cite{liu2024improved} and Qwen2VL~\cite{wang2024qwen2} as backbones. Extensive experiments on the MMEB benchmark consistently demonstrate significant performance gains across different base models, parameter scales, and learning resolutions. Notably, PDF boosts the VLM2Vec-LLaVA-1.6 (7B) model by a remarkable \textbf{+8.9} points on the low-resolution setting. While on the Qwen2VL backbone, it improves the 2B and 7B models by \textbf{+4.2} and \textbf{+3.1} points, respectively. Beyond sheer performance, PDF also demonstrates dramatic training efficiency. With only half the total computational budget, our 2B model surpasses the fully-computational baseline by \textbf{+2.6} points.

\section{Related Work}
\textbf{Multimodal Large Language Models} (MLLMs) have empowered LLMs with sophisticated visual understanding. A pioneering work, LLaVA~\cite{liu2023visual}, established a paradigm by projecting features from a pre-trained vision encoder (e.g., CLIP~\cite{radford2021learning}) into the LLM's word embedding space. Following this, the field has rapidly evolved, with research focusing on enhancing visual capabilities through dynamic high-resolution processing~\cite{liu2024llavanext} and scaling up vision encoders~\cite{chen2024internvl}, as seen in models like the Qwen-VL series~\cite{bai2023qwenvlversatilevisionlanguagemodel,wang2024qwen2}. This progress has opened up new avenues for applying these powerful generalist models to specialized domains, benefiting many downstream tasks, such as image classification~\cite{ju2023turbo,cheng2023category,cheng2023image}, image-text retrieval~\cite{wang2025advancing,lin2025squeeze,chen2023enhancing,cheng2023mixer,wang2025folder}, object segmentation~\cite{ye2021unsupervised,ma2025freesegdiff}, open vocabulary~\cite{ju2023multi,ma2024open,yang2024multi,ma2023open,yang2023multi,ma2023attrseg}, image AIGC~\cite{ma2023diffusionseg,chen2024wear,liu2023audio,ju2024turbo,yao2025beyond}, action recognition~\cite{zhao2020bottom,ju2022prompting,ju2022adaptive,ju2020point,cheng2024denoiser}, video grounding~\cite{ju2023constraint,liu2022exploiting,wang2025contrast,liu2024annotation,liu2024audio}, and temporal localization~\cite{ju2023distilling,ju2021divide}.

\textbf{Multimodal Embeddings without Large-Scale Models.} 
Before MLLMs, multimodal embeddings are dominated by dual-encoder architectures. Models such as CLIP~\cite{radford2021learning}, ALIGN~\cite{jia2021scaling}, and BLIP~\cite{li2022blip} trained separate image and text encoders via contrastive learning on large-scale image-text pairs, excelling at tasks like zero-shot classification and text-image retrieval. Subsequent work, such as UniIR~\cite{wei2024uniir}, sought to endow these dual-encoder models with more universal retrieval capabilities by introducing a comprehensive training dataset and the MBEIR benchmark. Despite the advances, their separated-encoder design inherently struggles with tasks requiring a deeper, holistic understanding or nuanced instruction-following.

\textbf{Multimodal Embeddings with Large-Scale Models.}
MLLMs' unified architecture naturally overcomes the limitations of dual-encoder models. Recent work explores to adapt MLLMs for embedding tasks, such as E5-V~\cite{jiang2024e5} and VLM2Vec~\cite{jiang2024vlm2vec} demonstrating state-of-the-art performance by fine-tuning MLLMs with contrastive losses on benchmarks like MMEB. LamRA~\cite{jiang2024vlm2vec} employs a retrieve-then-rerank pipeline, where an MLLM-based model reranks candidates from an initial retrieval stage. MMRet~\cite{zhou2024megapairs} introduced MegaPairs, a large-scale, instruction-tuning dataset for retrieval, and showed that pre-training on this data significantly gains downstream performance. Despite the success, these methods still adhere to a paradigm of \textit{single input, singular embedding, contrastive loss}, treating MLLM as a generator of a singular embedding for any given input. Our work challenges this ubiquitous paradigm through introducing a novel PDF framework, to explicitly explore the intrinsic semantic diversity within a single input by generating multiple, decoupled representations, one direction largely unexplored.

\section{PDF: Parallel Decoupling Framework}
\begin{figure*}[tp]
\centering
\includegraphics[width=\linewidth]{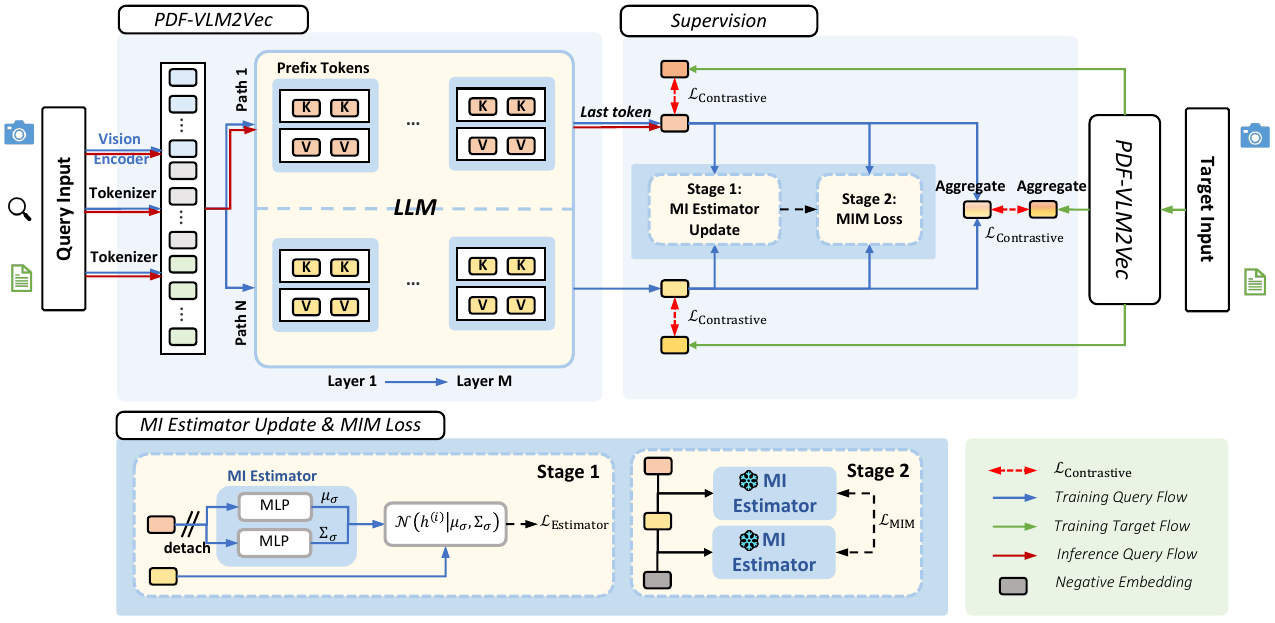}
\vspace{-18pt}
\caption{\textbf{Overview of the PDF-VLM2Vec Pipeline.} 
The training process (blue and green lines) is guided by a dual-objective system. For each input, parallel embeddings are generated via learnable prefixes. These are then supervised by: (1) Contrastive Loss to maintain representation quality, and (2) MIM Loss to enforce diversity. The MIM loss is calculated in a two-stage process: first updating the MI estimator with detached embeddings (Stage 1), and then using the frozen estimator to compute the loss (Stage 2). During inference (red line), a single forward pass inherits these benefits, yielding a robust embedding with additional negligible computational overhead.}
\label{fig:pipeline}
\vspace{-10pt}
\end{figure*}

Our \textbf{P}arallel \textbf{D}ecoupling \textbf{F}ramework \textbf{(PDF)} learns multiple parallel embeddings under the guidance of Mutual Information Minimization (MIM). Fig.~\ref{fig:pipeline} illustrates the framework overview. Our method operates on query-target pairs, denoted as $(q, t^+)$, where each element can be a single image, a single text, or one interleaved combination of both. We formally define them as $q = (q_t, q_i)$ and $t^+ = (t^+_t, t^+_i)$, where the subscripts $t$ and $i$ denote the text and image components, respectively. A component can be absent if the input is unimodal. Following prior work~\cite{jiang2024vlm2vec}, we augment the query $q$ with a task-specific instruction template to form the final input $q_{inst}$:
\begin{equation}
\label{eq:input_template}
q_{inst} = [\text{IMAGE\_TOKEN}] \text{ Instruct: } \{ \text{task\_definition} \} \text{ Query: } \{q\},
\end{equation}
where $\{\text{task\_definition}\}$ is a placeholder for a concise description of the embedding task. Note that we do not augment the target input with task instruction, following consensus.

During training, our PDF framework creates $N$ parallel computational paths within the LLM backbone, generating distinct embeddings for each query input $q_{inst}$ and the target input $t^+$, namely $\{h^{(i)}_{q}\}_{i=1}^N$ and $\{h^{(i)}_{t+}\}_{i=1}^N$. This is achieved by conditioning each path on a set of learnable ``prefix" parameters. These parallel embeddings, along with an aggregation module computed via lightweight MLP-softmax to generate the aggregated query embedding $h_{q}$ and aggregated target embedding $h_{t+}$. These embeddings are supervised by our comprehensive losses (Sec.~\ref{sec:loss}), which integrates a contrastive objective for alignment with a MIM objective for diversity.

At inference time, the parallel mechanism and aggregation module are bypassed for efficiency, as shown by the red line in Fig.~\ref{fig:pipeline}. We adopt only a single pre-determined path ({\em e.g.}, the one corresponding to the first prefix) to get final outputs, making the inference procedure identical to that of the baseline and introducing negligible additional computational overhead or latency. As a result, our method retains the benefits of enriched training without incurring any deployment cost.

\subsection{Generating Diverse Embeddings via Deep Prefix Injection}
To foster the exploration of a diverse embedding space, we introduce a deep prefix injection mechanism. Drawing inspiration from prefix-tuning~\cite{li2021prefix}, {\em i.e.}, models can be steered by conditioning on different prefixes, we apply the idea at a deeper level to create our $N$ parallel computational paths. Instead of merely prepending tokens to the input sequence, we inject path-specific, learnable prefix parameters directly into each transformer layer of the LLM backbone.

We define a unique prefix that modulates the self-attention mechanism within each of the $M$ transformer layers for each of the $N$ parallel paths. For path $i$ and layer $l$, a specific prefix is denoted as $(\mathbf{p}_\mathbf{K}^{(i,l)} ,\mathbf{p}_\mathbf{V}^{(i,l)}) \in (\mathbb{R}^{K\times d},\mathbb{R}^{K\times d})$, where $K$ is the length of prefix tokens and $d$ is the dimension.

Then we use prefixes in the self-attention mechanism in LLM. Specifically, for each layer of each path, we first project the token sequence to obtain $\mathbf{Q} \in \mathbb{R}^{L \times d}$, $\mathbf{K} \in \mathbb{R}^{L \times d}$ and $\mathbf{V} \in \mathbb{R}^{L \times d}$, where the $L$ is the sequence length. Then we concatenate the $\mathbf{K}$ and $\mathbf{V}$ with their corresponding prefix:
\begin{equation}
    \mathbf{K}' = \text{Concat}(\mathbf{p}_{K}^{(i,l)}, \mathbf{K})\in \mathbb{R}^{(L+K)\times d}, \quad \mathbf{V}' = \text{Concat}(\mathbf{p}_{V}^{(i,l)}, \mathbf{V})\in \mathbb{R}^{(L+K)\times d}.
    \label{eq:prefix_concat} 
\end{equation}
Then the $\mathbf{Q}$ performs attention over augmented key-value pairs. This layer-wise modulation, repeated across the model's depth with path-specific parameters, guides each path to produce a distinct output sequence. Finally, we select the last token as the output embedding $h^{(i)}$ for path $i$.

Having obtained the parallel embeddings, one critical question then arises: \textit{\textbf{how can we explicitly promote diversity?}} While one might hope that the distinct prefixes would naturally lead to diverse outputs~\cite{chen2025parallel}, we contend that an explicit supervisory signal is far more effective. Therefore, we employ a Mutual Information Minimization (MIM) objective to actively discourage statistical dependence among the parallel embeddings, thereby compelling the model to discover a multifaceted embedding space that covers comprehensive semantics.

\subsection{Promoting Diverse Embeddings Via Mutual Information Minimization}
\label{sec:method_mim}
To explicitly enforce diversity among the $N$ parallel embeddings $\{h^{(i)}\}_{i=1}^N$, our goal is to minimize the Mutual Information (MI) between any pair of them, $I(h^{(i)}; h^{(j)})$. However, directly computing MI is intractable as it requires access to the true data distributions. We hence minimize a tractable variational upper bound, formulated as the two-stage optimization.

\textbf{Variational Upper Bound.}
To minimize a tractable upper bound of MI, we adopt the Contrastive Log-ratio Upper-Bound (CLUB)~\cite{cheng2020club} estimator. CLUB introduces a variational distribution $q_\sigma(h^{(i)}|h^{(j)})$, parameterized by a neural network (\textit{\textbf{MI Estimator}}) with parameters $\sigma$, to approximate the true conditional probability $p(h^{(i)}|h^{(j)})$. MI is then upper-bounded by:
\begin{equation}
\label{eq:club_bound}
I(h^{(i)}; h^{(j)}) \le \mathbb{E}_{p(h^{(i)}, h^{(j)})}[\log q_\sigma(h^{(i)}|h^{(j)})] - \mathbb{E}_{p(h^{(i)})p(h^{(j)})}[\log q_\sigma(h^{(i)}|h^{(j)})].
\end{equation}
Minimizing the upper bound encourages embeddings $h^{(i)}$ and $h^{(j)}$ to be statistically independent.

\textbf{Two-Stage Adversarial-like Optimization.}
Within each training iteration, the process unfolds as a two-stage game, where the MI Estimator and the main MLLM play opposing roles.

$\bullet$ \underline{\textit{Training the MI Estimator.}} First, we train the MI Estimator to detect statistical dependencies between parallel embeddings. Given a pair $(h^{(i)}, h^{(j)})$ generated from the \textit{same} input sample (a ``positive'' pair), we train the estimator to predict $h^{(i)}$ from $h^{(j)}$ by maximizing the conditional log-likelihood. We parameterize $q_\sigma$ as a Gaussian, $q_\sigma(h^{(i)}|h^{(j)}) = \mathcal{N}(h^{(i)} | \mu_\sigma(h^{(j)}), \Sigma_\sigma(h^{(j)}))$, where an MLP predicts the mean $\mu$ and covariance $\Sigma$. The objective for the estimator's parameters $\sigma$ is:
\begin{equation}
\label{eq:estimator_loss}
\mathcal{L}_{\text{Estimator}} = - \mathbb{E}_{p(h^{(i)}, h^{(j)})}[\log q_\sigma(h^{(i)}|h^{(j)})].
\end{equation}
Crucially, during this stage, the gradients from this loss only update $\sigma$; the embeddings $h^{(i)}$ and $h^{(j)}$ are \textit{detached} from the computation graph of MLLMs.

$\bullet$ \underline{\textit{Training the MLLM backbone.}} Second, with the MI Estimator's parameters $\sigma$ \textit{frozen}, we update the MLLM parameters $\theta$. The goal now is to make the estimator's job harder by generating embeddings that are less predictable. We do this by minimizing the MI upper bound from Eq.~(\ref{eq:club_bound}), which serves as our MIM loss. For a mini-batch of size $B$, $\mathcal{L}_{\text{MIM}}$ is approximated as:
\begin{equation}
\label{eq:mim_loss}
\mathcal{L}_{\text{MIM}} = \frac{1}{B} \sum_{k=1}^{B} \frac{1}{N(N-1)} \sum_{i \neq j} \left( \log q_\sigma(h_k^{(i)} | h_k^{(j)}) - \mathbb{E}_{m \neq k}[\log q_\sigma(h_k^{(i)} | h_m^{(j)})] \right).
\end{equation}
Here, the first term is the log-likelihood for a ``positive'' pair from the same sample $k$. The second term is the expectation over ``negative'' pairs, approximated by pairing $h_k^{(i)}$ with an embedding $h_m^{(j)}$ from another sample $m$ within the same batch. The total gradient from $\mathcal{L}_{\text{MIM}}$ (and the contrastive loss) then updates $\theta$, pushing the parallel embeddings towards greater differentiation.

\subsection{Overall Training Objective}
\label{sec:loss}
Our training objective comprises two crucial components: one contrastive loss $\mathcal{L}_{\text{contrastive}}$ to enforce representation quality, and one Mutual Information Minimization loss $\mathcal{L}_{\text{MIM}}$ to promote diversity.

\textbf{Contrastive Loss.}
Our primary supervision signal is the InfoNCE loss~\cite{oord2018representation}, which enforces query-target semantic alignment. We first apply it to the aggregated embeddings: for a given query's aggregated embedding $h_{q}$, its corresponding aggregated target embedding $h_{t^+}$ serves as the positive sample, while all other target embeddings in the mini-batch $\mathcal{B}^{-}$ act as negatives:
\begin{equation}
\label{eq:infonce}
\mathcal{L}_{\text{InfoNCE}}(h_{q}, h_{t^+}) = -\log \frac{\exp(\text{sim}(h_{q}, h_{t^+}) / \tau)}{\exp(\text{sim}(h_{q}, h_{t^+}) / \tau) + \sum_{h_{t^-} \in \mathcal{B}^{-}} \exp(\text{sim}(h_{q}, h_{t^-}) / \tau)},
\end{equation}
where $\text{sim}(\cdot, \cdot)$ is cosine similarity and $\tau$ is a temperature hyperparameter.

To prevent the diversity encouraged by $\mathcal{L}_{\text{MIM}}$ from degrading representation quality, we also apply the InfoNCE loss to each of the $N$ parallel paths. This serves as a crucial representation constraint, compelling each path to learn semantically valid representations rather than collapsing into several trivial solutions ({\em e.g.}, random noise) to minimize mutual information. Therefore, the contrastive loss supervises both aggregated and parallel embeddings, by a weighting hyperparameter $\lambda_{\text{CON}}$. 
\begin{equation}
\label{eq:contrastive_loss}
\mathcal{L}_{\text{contrastive}} = \mathcal{L}_{\text{InfoNCE}}(h_{q, \text{agg}}, h_{t^+}) + \lambda_{\text{CON}} \frac{1}{N} \sum_{i=1}^{N} \mathcal{L}_{\text{InfoNCE}}(h_q^{(i)}, h_{t^+}^{(i)}),
\end{equation}
where $h_{t^+}^{(i)}$ is the $i$-th parallel embedding of the positive target.

\textbf{Total Training Objective.}
Finally, our PDF framework is optimized through a linear combination of the contrastive loss for quality and the MIM loss (from Sec.~\ref{sec:method_mim}) for diversity:
\begin{equation}
\label{eq:total_loss}
\mathcal{L}_{\text{total}} = \mathcal{L}_{\text{contrastive}} + \lambda_{\text{MIM}} \mathcal{L}_{\text{MIM}}.
\end{equation}

\section{Experiment}
\subsection{Datasets \& Metrics \& Implementations}
\textbf{Datasets \& Metrics.} Following VLM2Vec~\cite{jiang2024vlm2vec}, we train on the 20 in-distribution MMEB datasets covering 662K pairs across four meta-tasks: classification, VQA, multimodal retrieval, and visual grounding. The model is then evaluated on both 20 in-distribution and 16 out-of-distribution MMEB test sets. We report Precision@1 as metrics on each dataset, {\em i.e.}, the proportion of top-ranked candidates that are positive samples.

\textbf{Implementation Details.} 
We instantiate PDF upon the VLM2Vec~\cite{jiang2024vlm2vec}, with multiple backbones, model scales and resolutions. For LLaVA-1.6 (7B), we train both a low-resolution (LR, 334x334) and a high-resolution (HR, 1344x1344) variant, referred to PDF-VLM2Vec-LLaVA1.6-LR and PDF-VLM2Vec-LLaVA1.6. For Qwen2VL, we train HR versions for both 2B and 7B scales, as well as an LR version (LR, 128x128) for the 2B model. These are denoted as PDF-VLM2Vec-Qwen2VL (for HR) and PDF-VLM2Vec-Qwen2VL-LR (for the 2B LR). For efficient fine-tuning, we apply LoRA~\cite{hu2022lora} to the LLM backbone with a rank of $r=8$ in all experiments.

We set the number of parallel paths to $N=2$. Each path is conditioned by a deep prefix of length $K=20$ injected into each transformer layer. The aggregated representation is computed through one lightweight MLP, which consists of two linear layers with a SiLU activation function. Besides, for the loss weight hyperparameters, we set $\lambda_{\text{MIM}}=1 \times 10^{-4}$ and $\lambda_{\text{CON}}=1.0$. These hyperparameters are determined based on ablation studies detailed in Appendix~\ref{app:hyperparams}.

All models are trained with a global batch size of 1024 and an InfoNCE temperature of $\tau=0.02$. To support this large batch size, we leverage GradCache~\cite{jiang2024vlm2vec}, following the original VLM2Vec implementation. We use the AdamW optimizer~\cite{loshchilov2017decoupled} with a peak learning rate of $2 \times 10^{-5}$ for the 2B models and $5 \times 10^{-6}$ for the 7B model. Besides, the training schedule consists of 2000 steps, including a 100-step linear warm-up followed by a linear learning rate decay. All experiments were conducted on NVIDIA H100 GPUs.

\textbf{Baselines.} 
We evaluate our method against two main categories of baselines. Our primary and most direct competitors are the \textbf{VLM2Vec}~\cite{jiang2024vlm2vec} models, the state-of-the-art framework upon which our work is built. To ensure a fair and direct comparison, for each of our PDF-enhanced models, we train its corresponding VLM2Vec counterpart using the identical dataset, hyperparameters, and overall training procedure. This strictly controlled setup allows us to isolate the performance gains attributable specifically to our PDF framework. Secondly, to contextualize our results within the broader landscape of multimodal representation learning, we also compare against a wide range of established models. Following the evaluation protocol from VLM2Vec, this group includes both prominent LLM-based and non-LLM-based methods. Unless otherwise specified, all baseline results are taken directly from original papers. The only exception is the VLM2Vec-Qwen2VL-LR baseline, which we reproduced under the same controlled settings for a direct comparison.

\subsection{Main Results}
\begin{table*}[!t]
\centering
\addtolength{\tabcolsep}{-1pt}
\small 
\caption{\textbf{Comparisons with SOTA on the MMEB benchmark.} Our PDF‑VLM2Vec is evaluated against both non‑LLM and LLM‑based baselines. Scores are averaged per meta‑task and reported for In‑Distribution (IND), Out‑of‑Distribution (OOD), and Overall performance. The direct baselines, VLM2Vec, are shaded in gray; $\Delta$ denotes the absolute improvement over the direct baseline.
}
\begin{tabular}{l ccccccc}
\toprule
\multirow{2}{*}{\textbf{Model}} & \multicolumn{4}{c}{\textbf{Per Meta-Task Score}} & \multicolumn{3}{c}{\textbf{Average Score}} \\
\cmidrule(lr){2-5} \cmidrule(lr){6-8} 
& Class. & VQA & Retrieval & Grounding & IND & OOD & Overall \\
\midrule
\textbf{\# of datasets $\to$} & 10 & 10 & 12 & 4 & 20 & 16 & 36 \\
\midrule
\multicolumn{8}{c}{\textit{No LLM based Method}} \\
\midrule
CLIP~\cite{radford2021learning}                  & 42.8 & 9.1  & 53.0 & 51.8 & 37.1 & 38.7 & 37.8 \\
BLIP2~\cite{li2023blip}                  & 27.0 & 4.2  & 33.9 & 47.0 & 25.3 & 25.1 & 25.2 \\
SigLIP~\cite{zhai2023sigmoid}                & 40.3 & 8.4  & 31.6 & 59.5 & 32.3 & 38.0 & 34.8 \\
OpenCLIP~\cite{cherti2023reproducible}              & 47.8 & 10.9 & 52.3 & 53.3 & 39.3 & 40.2 & 39.7 \\
UniIR(CLIP\_SF)~\cite{wei2024uniir}        & 44.3 & 16.2 & 61.8 & 65.3 & 47.1 & 41.7 & 44.7 \\
CLIP-FFT~\cite{jiang2024vlm2vec}               & 55.2 & 19.7 & 53.2 & 62.2 & 47.6 & 42.8 & 45.4 \\
OpenCLIP-FFT~\cite{jiang2024vlm2vec}           & \textbf{56.0} & \textbf{21.9} & \textbf{65.4} & \textbf{64.1} & \textbf{50.5} & \textbf{43.1} & \textbf{47.2} \\
\midrule
\multicolumn{8}{c}{\textit{LLM-based model (2B model)}} \\
\midrule
ColPali v1.3~\cite{faysse2024colpali}         & 40.3 & 11.5 & 48.1 & 40.3 & -    & -    & 34.9 \\
GME~\cite{zhang2024gme}                 & 54.4 & 29.9 & 66.9 & 55.5 & -    & -    & 51.9 \\
\rowcolor{baselinegray}
VLM2Vec-Qwen2VL-LR   & 51.9 & 29.6 & 54.9 & 50.6 & 50.1 & 41.5 & 46.5 \\
PDF-VLM2Vec-Qwen2VL-LR      & 59.3 & 47.5 & 63.3 & 70.0 & 62.8 & 53.3 & 58.6 \\
\rowcolor{improvementblue}
\textbf{$\Delta$ - baseline}         & +7.4 & +17.9& +8.4 & +19.4& +12.7& +11.8& +12.1\\
\rowcolor{baselinegray}
VLM2Vec-Qwen2VL      & 58.7 & 49.3 & 65.0 & 72.9 & 65.6 & 52.3 & 59.7 \\
PDF-VLM2Vec-Qwen2VL         & \textbf{62.1} & \textbf{55.4} & \textbf{67.1} & \textbf{80.3} & \textbf{69.5} & \textbf{56.9} & \textbf{63.9} \\
\rowcolor{improvementblue}
\textbf{$\Delta$ - baseline}         & +3.4 & +6.1 & +2.1 & +7.4 & +3.9 & +4.6 & +4.2 \\
\midrule
\multicolumn{8}{c}{\textit{LLM-based model (7B model)}} \\
\midrule
GME~\cite{zhang2024gme}                 & 57.7 & 34.7 & \textbf{71.2} & 59.3 & -    & -    & 56.0 \\
LamRA-Qwen2~\cite{liu2025lamra}          & 59.2 & 26.5 & 70.0 & 62.7 & -    & -    & 54.1 \\
LamRA-Qwen2.5~\cite{liu2025lamra}        & 51.7 & 34.1 & 66.9 & 56.7 & -    & -    & 52.4 \\
\rowcolor{baselinegray}
VLM2Vec-LLAVA-1.6-LR     & 54.7 & 50.3 & 56.2 & 64.0 & 61.0 & 45.7 & 55.0 \\
PDF-VLM2Vec-LLAVA-1.6-LR         & 58.7 & 55.2 & 67.5 & 88.1 & 69.8 & 56.7 & 63.9 \\
\rowcolor{improvementblue}
\textbf{$\Delta$ - baseline}         & +4.0 & +4.9 & +11.3 & +24.1 & +8.8 & +11.0 & +8.9 \\

\rowcolor{baselinegray}
VLM2Vec-LLAVA-1.6     & 61.2 & 49.9 & 67.4 & 86.1 & 67.5 & 57.1 & 62.9 \\
PDF-VLM2Vec-LLAVA-1.6         & 59.7 & 56.1 & 67.8 & \textbf{89.2} & 70.4 & 57.5 & 64.7 \\
\rowcolor{improvementblue}
\textbf{$\Delta$ - baseline}         &-1.5  & +6.2 & +0.4 & +3.1 & +2.9 & +0.4 & +1.8 \\

\rowcolor{baselinegray}
VLM2Vec-Qwen2VL     & 62.7 & 56.9 & 69.4 & 82.2 & 71.9 & 57.5 & 65.5 \\
PDF-VLM2Vec-Qwen2VL          & \textbf{65.4} & \textbf{63.0} & 70.0 & 86.5 & \textbf{74.0} & \textbf{61.8} & \textbf{68.6} \\
\rowcolor{improvementblue}
\textbf{$\Delta$ - baseline}         & +2.7 & +6.1 & +0.6 & +4.3 & +2.1 & +4.3 & +3.1 \\
\bottomrule
\end{tabular}
\label{tab:main_results}
\vspace{-14pt}
\end{table*}

$\bullet$\,\underline{\textit{Comparisons with SOTA on the MMEB benchmark.}}
Table~\ref{tab:main_results} summarizes the performance of our PDF-VLM2Vec against various baselines. The direct baseline, VLM2Vec-Qwen2VL, is shaded in gray, while the absolute gains ($\Delta$) of our method are highlighted in blue.

The results unequivocally demonstrate the broad effectiveness and generalizability of our PDF training strategy. Our method consistently outperforms the VLM2Vec baselines, across different model scales, data resolutions and foundational MLLM. For the primary Qwen2VL baseline, PDF-VLM2Vec achieves substantial overall improvements of \textbf{+12.1}, \textbf{+4.2}, and \textbf{+3.1} points for the low-res 2B, high-res 2B, and high-res 7B models, respectively. The versatility of our strategy is further evidenced by its application to VLM2Vec-LLaVA-1.6 (7B), where it delivers impressive gains of \textbf{+8.9} points in the low-resolution setting and \textbf{+1.8} points in the high-resolution setting. This consistent performance enhancement across two distinct VLM architectures strongly validates the scalability and universal applicability of our proposed method.

$\bullet$\,\underline{\textit{Zero-Shot Image/Text Retrieval.}}
To further validate generalization, we conduct zero-shot retrieval experiments on three unseen datasets: Flickr30K~\cite{plummer2015flickr30k}, ShareGPT4V~\cite{chen2024sharegpt4v}, and Urban1K~\cite{zhang2024long}. As shown in Table~\ref{tab:zero_shot}, our PDF framework delivers compelling performance gains. The improvements are particularly dramatic for the 2B model, which sees performance boosts of up to \textbf{+17.2} points on Urban1K. Moreover, our method consistently improves upon the already strong 7B baseline across all benchmarks. These results demonstrate that our approach not only substantially enhances the generalization of smaller models but also robustly scales to larger, more capable ones, confirming its broad effectiveness.

\begin{table*}[!t]
    \centering
    \small
     \renewcommand{\arraystretch}{0.95}
    \addtolength{\tabcolsep}{-3pt}
    \label{tab:zero_shot}
    \caption{
    \textbf{Zero-shot image-text retrieval on the unseen datasets: Flickr30K, ShareGPT4V and Urban1K.} 
    Recall@1 (R@1) scores are reported. For both 2B and 7B model scales, $\Delta$ denotes the absolute point improvement over the direct baseline.
}
    \begin{tabular}{@{}lcccccc@{}}
        \toprule
        \multirow{2}{*}{Model} & \multicolumn{3}{c}{Text - Image (R@1)} & \multicolumn{3}{c}{Image - Text (R@1)} \\
        \cmidrule(lr){2-4} \cmidrule(lr){5-7}
                               & Flickr30K & ShareGPT4V & Urban1K & Flickr30K & ShareGPT4V & Urban1K \\
        \midrule
        CLIP                   & 79.5      & 90.1       & 77.8    & 92.9      & 93.6       & 80.7    \\
        EVA-CLIP-8B            & 80.3      & 93.1       & 80.4    & 94.5      & 91.2       & 77.8    \\
        E5-V (7B)              & 77.3      & 85.1       & 88.9    & 85.7      & 82.1       & 83.2    \\
        LAMRA-RET (7B)         & 82.8      & 93.3       & 95.1    & 92.7      & 88.1       & 94.3    \\
        \midrule
                \rowcolor{baselinegray}
        VLM2Vec-LLAVA1.6 (7B)   & 76.0      & 85.8       &  84.7   & 90.6      &  90.7      &   90.8  \\
        PDF-VLM2Vec-LLAVA1.6 (7B)        & 80.0      &  87.8      &  90.5   &  93.9       &  93.5      &   91.7  \\
        \rowcolor{improvementblue}
        \textbf{$\Delta$ - baseline}     & +4.0     &  +2.0      &   +5.8  &   +3.3   &   +2.8    & +0.9    \\
        \midrule
        \rowcolor{baselinegray}
        VLM2Vec-Qwen2VL (2B)   & 68.4      & 89.4       & 75.5    & 84.5      & 92.1       & 84.4    \\
        PDF-VLM2Vec-Qwen2VL (2B)        & 77.1      & 94.7       & 92.7    & 92.0        & 95.8       & 93.6    \\
        \rowcolor{improvementblue}
        \textbf{$\Delta$ - baseline}              & +8.7      & +5.3       & +17.2   & +7.5      & +3.7       & +9.2    \\
        \midrule
        \rowcolor{baselinegray}
        VLM2Vec-Qwen2VL (7B)   & 79.6      & 93.0         & 92.9    & 91.6      & 96.7       & 93.2    \\
        PDF-VLM2Vec-Qwen2VL (7B)        & 79.7      & 95.3       & 96      & 92.3      & 97.4       & 95.9    \\
        \rowcolor{improvementblue}
        \textbf{$\Delta$ - baseline}             & +0.1      & +2.3       & +3.1    & +0.7      & +0.7       & +2.7    \\
        \bottomrule
    \end{tabular}
\end{table*}

\subsection{Ablation Study}
\begin{table}[t!]
\centering
\small
\addtolength{\tabcolsep}{1.5pt}
\caption{\textbf{Ablation study of all components.} We start from the baseline (R1) and incrementally add deep prefix tunning (Prefix, R2), Parallel Embeddings (Parallel, R3), MIM Loss (MIM, R4), and Subspace Contrastive Loss (Sub Loss, R5). R6-R7 compares alternative inference strategies.}
\begin{tabular}{l ccccc ccc}
\toprule
\multirow{2}{*}{\textbf{ID}} & \multicolumn{5}{c}{\textbf{Module}} & \multicolumn{3}{c}{\textbf{Average Score}} \\
\cmidrule(lr){2-6} \cmidrule(lr){7-9}
& Prefix & Parallel Path & MIM & Sub Loss & Inference Strategy & IND & OOD & Overall \\
\midrule
R1 & - & - & - & - & -  & 65.6 & 52.3 & 59.7 \\
R2 & \checkmark & - & - & - & Single Prefix  & 64.9 & 53.2 & 59.7 \\
R3 & \checkmark & \checkmark & - &- & Single Prefix  & 67.2 & 55.8 & 62.1 \\
R4  & \checkmark & \checkmark & \checkmark & - & Single Prefix  & 68.0 & \textbf{57.4} & \underline{63.3} \\
\textbf{R5} & \checkmark & \checkmark & \checkmark & \checkmark  & Single Prefix & \textbf{69.6} & \underline{56.9} & \textbf{63.9} \\
\midrule
R6 & \checkmark & \checkmark & \checkmark & \checkmark& Aggregate   & \underline{69.5} & \underline{56.9} & \textbf{63.9} \\
R7 & \checkmark & \checkmark & \checkmark & \checkmark & No Prefix & 33.4 & 25.0 & 29.7 \\
\bottomrule
\end{tabular}
\label{tab:ablation_study}
\end{table}

Table~\ref{tab:ablation_study} conducts comprehensive ablation studies, structured in two parts: validating the effectiveness of components and evaluating different inference strategies. All experiments are performed on the high-resolution MMEB dataset using VLM2Vec-Qwen2VL (2B).

$\bullet$\,\underline{\textit{Effectiveness of Training Components.}}
The top section of the table illustrates the step-by-step construction of our model. We begin with the VLM2Vec baseline (R1, 59.7 Overall). First, to verify that our gains do not simply stem from increased parameterization, we add prefix parameters to the baseline in a single-path setting (R2). This yields no performance improvement, confirming that the parallel architecture is the true source of gains. Indeed, introducing \textbf{Parallel Paths} (R3) provides a significant +2.4 point uplift. Subsequently, applying the \textbf{MIM Loss} (R4) to encourage diversity brings a further +1.2 point improvement. Finally, adding the \textbf{Sub Loss} to enforce quality on each path (R5) results in our full model, achieving the best performance of 63.9 Overall. This incremental improvement at each step validates the efficacy of our core design choices.

$\bullet$\,\underline{\textit{Analysis of Inference Strategies.}} 
The bottom section of the table investigates different inference strategies, all applied to the same fully-trained model (R5). Most critically, removing the prefix during inference (R7) leads to a catastrophic performance collapse (29.7 Overall), demonstrating that the learned prefixes are essential for activating the correct representational subspaces. Furthermore, we observe that using the aggregated embedding (R6) yields identical performance to our proposed strategy of using a single prefix path (R5). Given that the single-path approach incurs almost the same computation overload compared to the baseline (Table~\ref{tab:inference_efficiency}), which is ideal for deployment.

\subsection{Analysis of Efficiency and Diversity}
\begin{table*}[h!]
\centering
\small
\caption{\textbf{Inference efficiency of PDF.} 
We analyze the parameter and computational overhead of PDF on the VLM2Vec-Qwen2VL (2B) model. Inference TFLOPs are calculated using one 1344x1344 image. The table highlights that our standard inference strategy (``Single Prefix") achieves a significant +4.2 point performance gain over the baseline with negligible additional computational overhead (\textbf{+0.06\%}) and only one minor increase in trainable parameters.}
\label{tab:inference_efficiency}
\begin{tabular}{l l c c c c}
\toprule
\textbf{Model} & \textbf{Inference} & \textbf{Params (B)} & \textbf{Trainable (\%) } & \textbf{TFLOPs} & \textbf{MMEB Score} \\
\midrule
VLM2Vec (LoRA, r=8)  & -                       & 2.214      & 0.211                   & 18.925                    & 59.7 \\
VLM2Vec (LoRA, r=16) & -                       & 2.218         & 0.415                   & 18.925                    & 56.0 \\
\midrule
\multirow{2}{*}{PDF-VLM2Vec} & Aggregate       & \multirow{2}{*}{2.219} & \multirow{2}{*}{0.449} & 24.999 & 63.9 \\
 & \textbf{Single Prefix}  &                       &                          & \textbf{18.937}           &  \textbf{63.9}    \\
\bottomrule
\end{tabular}

\vspace{-10pt}

\end{table*}

\begin{table*}[h!]
\centering
\small
\caption{\textbf{Training efficiency of PDF.} Our PDF-VLM2Vec demonstrates significant improvements in the training efficiency compared to the VLM2Vec baseline.}
\begin{tabular}{ll ccc ccc}
\toprule
\multirow{2}{*}{\textbf{Model}} & \multirow{2}{*}{\textbf{Iters}} & \multicolumn{3}{c}{\textbf{2B Model}} & \multicolumn{3}{c}{\textbf{7B Model}} \\
\cmidrule(lr){3-5} \cmidrule(lr){6-8}
 & & \textbf{IND} & \textbf{OOD} & \textbf{Overall} & \textbf{IND} & \textbf{OOD} & \textbf{Overall} \\
\midrule
VLM2Vec-Qwen2VL & 2000 & 65.6 & 52.3 & 59.7 & 71.9 & 57.5 & 65.5 \\
\midrule
\multirow{3}{*}{PDF-VLM2Vec-Qwen2VL}
 & 500  & 66.4 & 57.3 & 62.3 \deltares{+2.6} & 70.8 & 61.2 & 66.5 \deltares{+1.0} \\
 & 1000 & 68.3 & \textbf{57.5} & 63.5 \deltares{+3.8} & 72.8 & 60.2 & 67.1 \deltares{+1.6} \\
 & 2000 & \textbf{69.5} & 56.9 & \textbf{63.9} \deltares{+4.2} & \textbf{74.0} & \textbf{61.8} & \textbf{68.5} \deltares{+3.0} \\
\bottomrule
\end{tabular}
\label{tab:training_efficiency}
\vspace{-10pt}
\end{table*}

We conduct a comprehensive analysis to demonstrate that PDF's performance gains are achieved with remarkable efficiency, both in terms of computation and parameters.

$\bullet$\,\underline{\textit{Parameter \& Inference Efficiency.}}
As detailed in Table~\ref{tab:inference_efficiency}, the performance gains are not a byproduct of increased parameters. Baseline VLM2Vec with a doubled LoRA rank (r=16) actually shows a performance drop, whereas our PDF, with a comparable number of trainable parameters (0.449\% {\em vs.} 0.415\%), delivers one substantial +7.9\% improvement (63.9 {\em vs.} 56.0). Most critically, our standard ``Single Prefix" inference incurs virtually no additional computational cost (18.937 {\em vs.} 18.925 TFLOPs), confirming that the benefits of our diversified training are inherited at zero overhead.

\begin{wrapfigure}{r}{0.5\linewidth} 
    \centering
    \includegraphics[width=\linewidth]{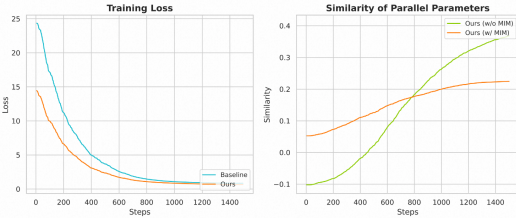}
\caption{
\textbf{Record of the training dynamics on VLM2Vec-Qwen2VL (2B).} 
\textbf{Left}: Training loss {\em vs.} baseline. 
\textbf{Right}: Cosine similarity of parallel prefixes w/ and w/o MIM loss.
}
    \label{fig:loss_similarity}
\end{wrapfigure}

$\bullet$\,\underline{\textit{Training Efficiency \& Convergence.}}
While parallel training of PDF increases computation per iteration (approximately 1.32 times, comparing Aggregate {\em vs.} baseline TFLOPs), this is overwhelmingly compensated by accelerated convergence. As shown in Fig.~\ref{fig:loss_similarity} Left, our 2B model surpasses the final performance of the fully-trained baseline (59.7) in just \textbf{500 iterations}, achieving a score of 62.3 as shown in Table~\ref{tab:training_efficiency}. This indicates our model reaches one superior state using only \textbf{50\% of the total computational budget} (500 iters\,$\times$\,2 paths {\em vs.}\,2000 iters\,$\times$\,1 path), highlighting a more effective learning process.

\vspace{0.3cm}
$\bullet$\,\underline{\textit{Visualizing Diversity.}}
Fig.~\ref{fig:loss_similarity} Right reveals the mechanism behind this efficiency. Without MIM constraint, these parallel prefixes' embeddings quickly collapse to similar. In contrast, this collapse is effectively counteracted with the MIM loss, enforcing diversity throughout training. We hypothesize this enforced diversity compels the model to explore a wider representational space, avoiding suboptimal local minima and accelerating the discovery of a robust, generalizable solution.

\section{Conclusion}
This paper introduces PDF, a novel VLM2Vec training framework designed to enhance the robustness and efficiency of embedding learning. Our core idea is to guide MLLMs to encourage diverse embedding exploration through parallel paths supervised by a Mutual Information Minimization (MIM) objective. During inference, we find the embedding space can be effectively activated by a single path, resulting in significant performance gains with negligible additional computational overhead. Extensive experiments on the MMEB benchmark and zero-shot benchmark validate the effectiveness across various tasks and model setting. We demonstrated that integrating PDF into the VLM2Vec framework not only substantially improves performance across a wide range of tasks and model scales but also accelerates training convergence. We hope that this work can inspire researchers to explore more effective training frameworks for the MLLM-based embedding models.

\bibliography{iclr2026_conference}
\bibliographystyle{iclr2026_conference}

\appendix

\newpage

\section{Appendix}
\subsection{Hyperparameter experiment on MMEB.}
\label{app:hyperparams}

Our PDF framework is governed by three primary hyperparameters: the number of parallel paths ($N$), the prefix length ($K$), and the MIM loss weight ($\lambda_\text{MIM}$). To efficiently determine an ideal configuration, we conduct a sensitivity analysis on the PDF-VLM2Vec-Qwen2VL (2B) model in a low-resolution setting on MMEB. Our search is centered around the configuration of $N=2$, $K=20$, and $\lambda_\text{MIM}=1 \times 10^{-4}$, while other hyperparameters ({\em e.g.}, learning rate) are kept consistent with our main experimental setup. The detailed results are presented in Table~\ref{tab:para}.

As shown in Table~\ref{tab:para}, the configuration of $N=2$, $K=20$, and $\lambda_\text{MIM}=1 \times 10^{-4}$ achieves the best overall performance. Notably, while the $N=4$ setting yields a slightly higher in-distribution (IND) score, it does so at the cost of doubling the training computation. Given this compelling trade-off between performance and efficiency, we adopt $N=2$ for all main experiments. This choice is further validated by our main results, which demonstrate that the hyperparameters selected in this simplified setting (low-resolution, 2B model) generalize effectively to higher-resolution inputs and the larger 7B model.
\begin{table}[h!]
\centering
\small
\caption{\textbf{Hyperparameter Analysis on MMEB.} 
Ablation study on the number of parallel paths ($N$), prefix length ($K$), and MIM loss weight ($\lambda_\text{MIM}$) using the PDF-VLM2Vec-Qwen2VL-LR model. The results show that the configuration of $N=2$, $K=20$, and $\lambda_\text{MIM}=10^{-4}$ achieves the optimal trade-off between performance and efficiency.}
\label{tab:para}
\begin{tabular}{@{}ccccc@{}}
\toprule
\multirow{2}{*}{Hyperparameter}
 & \multirow{2}{*}{Value} & \multicolumn{3}{c}{Average Score} \\
\cmidrule(l){3-5}
&   & IND & OOD & Overall \\
\midrule
\multirow{2}{*}{$N$} & \textbf{2} & 62.8 & \textbf{53.3} & \textbf{58.6} \\
  & 4 & \textbf{63.0} & 53.0 & \textbf{58.6} \\
\midrule
\multirow{3}{*}{$K$} & 10 & 62.4 & 51.3 & 57.4 \\
& \textbf{20} & \textbf{62.8} & \textbf{53.3} & \textbf{58.6} \\
& 40 & 61.9 & 52.3 & 57.7 \\
\midrule
\multirow{3}{*}{$\lambda_\text{MIM}$} & $1 \times 10^{-3}$ & 62.5 & 50.7 & 57.3 \\
 & $\mathbf{1 \times 10^{-4}}$ & \textbf{62.8} & \textbf{53.3} & \textbf{58.6} \\
& $1 \times 10^{-5}$ & 61.9 & 50.1 & 56.7 \\
\bottomrule
\end{tabular}
\end{table}

\subsection{Details of MI Estimator and Aggregation Module}
\label{app:module_details}

\textbf{MI Estimator.}
Our Mutual Information (MI) estimator is designed to approximate the conditional distribution $q(y|x)$ as a diagonal Gaussian, $\mathcal{N}(\mu, \Sigma)$. To achieve this, it employs two separate Multi-Layer Perceptrons (MLPs) to predict the mean ($\mu$) and the log-variance ($\log \sigma^2$, the diagonal of $\Sigma$).
\begin{itemize}
    \item \textbf{Mean Prediction Network ($\mu$):} This network takes an input of dimension $d_x$, projects it to an intermediate hidden dimension of $d_h/2$, applies a ReLU activation, and then projects it back to the output dimension $d_y$. Its structure is: \texttt{Linear($d_x$, $d_h/2$)} $\rightarrow$ \texttt{ReLU} $\rightarrow$ \texttt{Linear($d_h/2$, $d_y$)}.
    \item \textbf{Log-Variance Prediction Network ($\log \sigma^2$):} This network shares an identical architecture with the mean prediction network but includes an additional Tanh activation function at the end. This final Tanh layer serves to constrain the output values, enhancing training stability. Its structure is: \texttt{Linear($d_x$, $d_h/2$)} $\rightarrow$ \texttt{ReLU} $\rightarrow$ \texttt{Linear($d_h/2$, $d_y$)} $\rightarrow$ \texttt{Tanh}.
\end{itemize}
In our experiments, the input/output dimensions ($d_x, d_y$) are equal to the model's hidden size $d$, and the intermediate hidden dimension $d_h$ is set to $4d$.

\vspace{1em} 

\textbf{Aggregation Module.}
The aggregation module is a lightweight network designed to compute the fusion weights for the $N$ parallel embeddings. It first concatenates the $N$ embeddings, each of dimension $d$, into a single vector of size $N \times d$. This vector is then processed by an MLP followed by a Softmax layer.
\begin{itemize}
    \item \textbf{MLP:} The MLP consists of two linear layers with a Swish (SiLU) activation function in between. It maps the concatenated input vector from $N \times d$ to an intermediate dimension $d$, and then down to a vector of size $N$. The structure is: \texttt{Linear($N \times d$, $d$)} $\rightarrow$ \texttt{SiLU} $\rightarrow$ \texttt{Linear($d$, $N$)}.
    \item \textbf{Softmax:} A Softmax function is applied to the final $N$-dimensional output of the MLP to produce a set of normalized weights, which are then used for the weighted average of the parallel embeddings.
\end{itemize}
\subsection{Qualitative results}

To further verify the effectiveness of our PDF framework, we conduct a visual comparison between the VLM2Vec-Qwen2VL (7B) baseline and our PDF-enhanced model on the MMEB dataset.

\begin{figure}[tp]
    \centering
    \includegraphics[width=\linewidth]{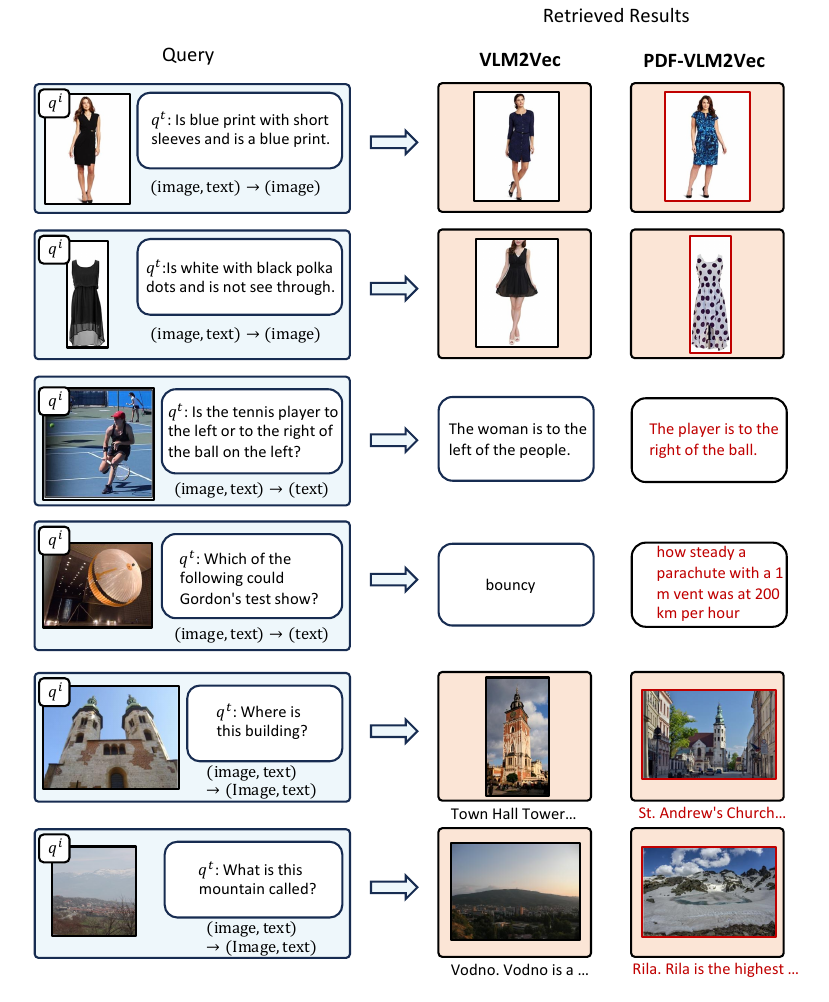}
    \caption{\textbf{Qualitative Results Part 1.} We show the results of our method across six different retrieval tasks compared with VLM2Vec-Qwen2VL-7B.}
    \label{fig:qualitative_1}
\end{figure}

\begin{figure}[tp]
    \centering
    \includegraphics[width=\linewidth]{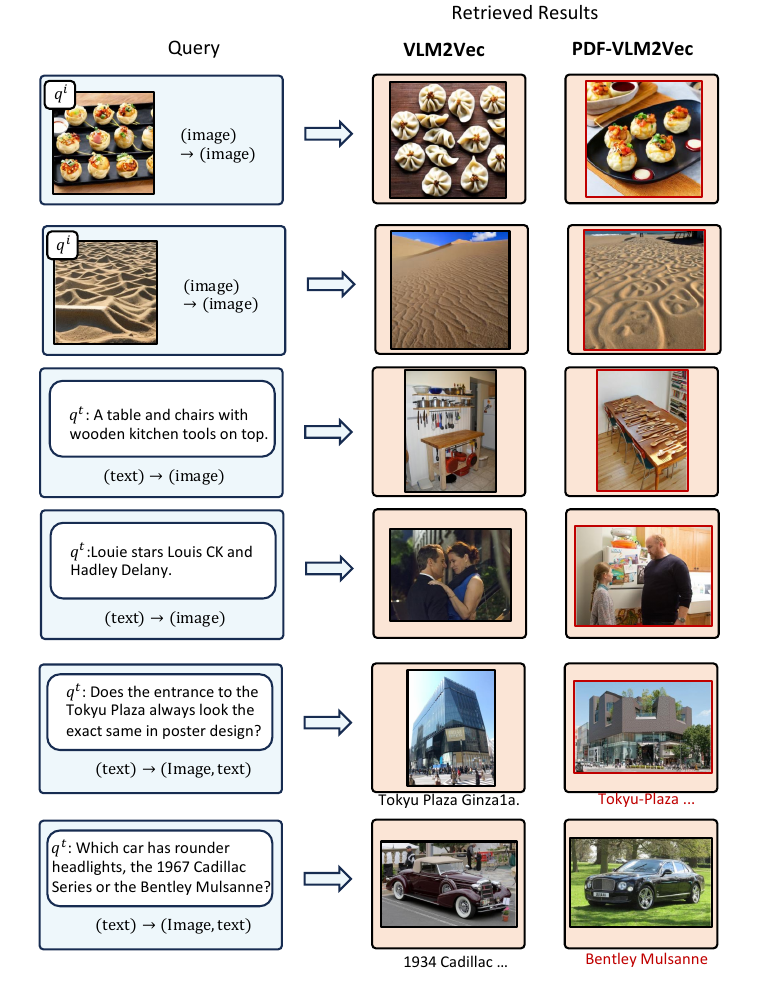}
    \caption{\textbf{Qualitative Results Part 2.} We show the results of our method across six different retrieval tasks compared with VLM2Vec-Qwen2VL-7B.}
    \label{fig:qualitative_2}
\end{figure}

As illustrated in Fig.~\ref{fig:qualitative_1} and Fig.~\ref{fig:qualitative_2}, including various multi-modal retrieval tasks, our PDF-enhanced model demonstrates a superior understanding of complex multi-modal queries, which require the model to retrieve target matching the requirement of image and text contents. As shown in the third row of Fig.~\ref{fig:qualitative_1}, the retrieved text of VLM2Vec can correctly matches the image content but neglects the crucial relationship specified in the query text. In contrast, our PDF-VLM2Vec successfully distinguishes this subtle difference and retrieves the correct result.

We attribute this improvement to our PDF, which compels the model to explore multifaceted features. Hence, our model trained with PDF can distinguish targets based on a wider range of aspects, leading to more precise and robust retrieval.

\subsection{Detail Results of the baseline and our VLM2Vec on MMEB}

We present the detailed results of each model on various datasets of MMEB in Table~\ref{tab:detailed_mmeb_results}.

\begin{table*}[!ht]
    \centering
    \small
    \addtolength{\tabcolsep}{-3.5pt}
    \caption{\textbf{Detailed Results on MMEB.} Detailed per-task performance comparison on the MMEB benchmark across all 36 tasks. The table reports Precision@1 for each task. Tasks marked with a highlighted background are from the out-of-distribution (OOD) test sets. \textbf{L} and \textbf{Q} means the LLAVA-1.6 and Qwen2VL backbones respectively.}
    \label{tab:detailed_mmeb_results}
    \begin{tabular}{l ccccccccccc}
        \toprule

        \multirow{2}{*}{\textbf{Task}}& \multirow{2}{*}{\textbf{CLIP}} & \multirow{2}{*}{\textbf{OpenCLIP}} & \multirow{2}{*}{\textbf{SigLIP}} & \multirow{2}{*}{\textbf{BLIP2}} & \multirow{2}{*}{\textbf{UniIR}} & \multicolumn{3}{c}{\textbf{VLM2Vec}} & \multicolumn{3}{c}{\textbf{PDF-VLM2Vec}} \\
        \cmidrule(lr){7-9} \cmidrule(lr){10-12}
         & & & & & &\textbf{L-7B} & \textbf{Q-2B} & \textbf{Q-7B} & \textbf{L-7B}& \textbf{Q-2B} & \textbf{Q-7B} \\
        \midrule

        \rowcolor{classcolor}
        \multicolumn{12}{l}{\textbf{Classification (10 tasks)}} \\
        ImageNet-1K & 55.8 & 63.5 & 45.4 & 10.3 & 58.3 &74.5 & 77.2 & 81.2 & 73.2 & 79.4 & 80.3\\
        N24News & 34.7 & 38.6 & 13.9 & 36.0 & 42.5 & 80.3 & 76.2 & 79.4& 80.1 & 78.6 & 82.5\\
        HatefulMemes & 51.1 & 51.7 & 47.2 & 49.6 & 56.4& 67.9 & 61.5 & 67.7 & 64.6 & 64.6  & 70.3\\
        VOC2007 & 50.7 & 52.4 & 64.3 & 52.1 & 66.2&91.5 & 79.0 & 81.7 & 89.6 & 82.7  & 84.3\\
        SUN397 & 43.4 & 68.8 & 39.6 & 34.5  & 63.2 &75.8 & 73.9 & 79.2 & 76.4 & 75.0  & 78.9\\
        \rowcolor{oodcolor}
        Place365 & 28.5 & 37.8 & 20.0 & 21.5 & 36.5 &44.0 & 36.0  & 38.4& 43.4 & 41.4 & 42.0\\
        \rowcolor{oodcolor}
        ImageNet-A & 25.5 & 14.2 & 42.6 & 3.2 & 9.8 & 43.6 & 51.5 & 55.3 & 43.0 & 53.8  & 53.9\\
        \rowcolor{oodcolor}
        ImageNet-R & 75.6 & 83.0 & 75.0 & 39.7  & 66.2 &79.8 & 86.4 & 74.5 & 77.9 & 89.2  & 85.7\\
        \rowcolor{oodcolor}
        ObjectNet & 43.4 & 51.4 & 40.3 & 20.6 & 32.2 &39.6 & 22.5 & 38.2 & 35.8 & 28.6  & 45.8\\
        \rowcolor{oodcolor}
        Country-211 & 19.2 & 16.8 & 14.2 & 2.5 & 11.3 &14.7 & 22.3 & 31.0 & 12.9 & 27.2 & 30.3\\
        \textit{All Classification} & 42.8 & 47.8 & 40.3 & 27.0 & 44.3&61.2 & 58.7& 62.7  & 59.7 & 62.1 & 65.4\\
        \midrule
        
        \rowcolor{vqacolor}
        \multicolumn{12}{l}{\textbf{VQA (10 tasks)}} \\
        OK-VQA & 7.5 & 11.5 & 2.4 & 8.7  & 25.4 &69.0 & 48.1 & 57.2 & 70.3 & 58.4 & 67.4\\
        A-OKVQA & 3.8 & 3.3 & 1.5 & 3.2  & 8.8 &54.4 & 40.3 & 48.0 & 58.0 & 51.2 & 59.1\\
        DocVQA & 4.0 & 5.3 & 4.2 & 2.6  & 6.2 &52.0 & 85.2  & 90.0 &80.4 & 89.5 & 92.4\\
        InfographicsVQA & 4.6 & 4.6 & 2.7 & 2.0  & 4.6 & 30.7& 49.3 & 65.0 & 42.0 & 56.5  & 68.1\\
        ChartQA & 1.4 & 1.5 & 3.0 & 0.5  & 1.6&34.8  & 42.0 & 55.3 & 45.7 & 50.0 & 60.2\\
        Visual7W & 4.0 & 2.6 & 1.2 & 1.3  & 14.5 & 49.8& 50.1& 53.0 & 52.5 & 52.6  & 54.6\\
        \rowcolor{oodcolor}
        ScienceQA & 9.4 & 10.2 & 7.9 & 6.8 & 12.8 &42.1 & 29.2 & 39.5 & 40.5 & 36.4  & 43.0\\
        \rowcolor{oodcolor}
        VizWiz & 8.2 & 6.6 & 2.3 & 4.0 & 24.3 &43.0 & 37.0 & 38.5 & 45.1 & 43.8  & 46.6\\
        \rowcolor{oodcolor}
        GQA & 41.3 & 52.5 & 57.5 & 9.7  & 48.8 & 61.2& 47.9 & 52.7 & 53.3 & 42.0  & 56.4\\
        \rowcolor{oodcolor}
        TextVQA & 7.0 & 10.9 & 1.0 & 3.3 & 15.1 & 62.0& 63.7& 70.2 & 73.6 & 73.7  & 81.7\\
        \textit{All VQA} & 9.1 & 10.9 & 8.4 & 4.2 & 16.2 & 49.9 & 49.3 & 56.9 &56.1 & 55.4  & 63.0\\
        \midrule
        
        \rowcolor{retcolor}
        \multicolumn{12}{l}{\textbf{Retrieval (12 tasks)}} \\
        VisDial & 30.7 & 25.4 & 21.5 & 18.0  & 42.2 & 80.9 & 75.5 & 81.3 & 82.7 & 80.0  & 82.7\\
        CIRR & 12.6 & 15.4 & 15.1 & 9.8 & 51.3 &49.9 & 48.5 & 50.0 & 51.8 & 51.6 & 53.0\\
        VisualNews\_t2i & 78.9 & 74.0 & 51.0 & 48.1  & 74.3 & 75.4 & 74.5 & 80.2& 74.9 & 74.3  & 80.0\\
        VisualNews\_i2t & 79.6 & 78.0 & 52.4 & 13.5 & 76.8 & 80.0 & 74.5 & 82.4 & 78.0 & 76.6  & 82.4\\
        MSCOCO\_t2i & 59.5 & 63.6 & 58.3 & 53.7  & 68.5 & 75.7 & 71.2& 77.2 & 76.9 & 73.1  & 76.9\\
        MSCOCO\_i2t & 57.7 & 62.1 & 55.0 & 20.3 & 72.1 & 73.1 & 68.2 & 73.2& 72.7 & 69.6  & 73.8\\
        NIGHTS & 60.4 & 66.1 & 62.9 & 56.5 & 66.2 & 65.5 & 65.1& 67.9 & 67.2 & 68.0  & 68.1\\
        WebQA & 67.5 & 62.1 & 58.1 & 55.4 & 89.6 &87.6 & 86.1 & 88.1 & 89.1 & 87.7  & 87.8\\
        \rowcolor{oodcolor}
        FashionIQ & 11.4 & 13.8 & 20.1 & 9.3 & 40.2 & 16.2 & 13.5 & 16.8 & 16.1 & 15.7  & 16.8\\
        \rowcolor{oodcolor}
        Wiki-SS-NQ & 55.0 & 44.6 & 55.1 & 28.7 & 12.2 & 60.2 & 57.7 & 61.4 & 67.0 & 58.6  & 65.6\\
        \rowcolor{oodcolor}
        OVEN & 41.1 & 45.0 & 56.0 & 39.5 & 69.4 &56.5 & 64.5& 67.4& 49.1 & 68.3  & 70.8\\
        \rowcolor{oodcolor}
        EDIS & 81.0 & 77.5 & 23.6 & 54.4 & 79.2& 87.8 & 80.1 & 87.1& 88.5 & 81.2  & 81.8\\
        \textit{All Retrieval} & 53.0 & 52.3 & 31.6 & 33.9 & 61.8 & 67.4 & 65.0 & 69.4 & 67.8 & 67.1  & 70.0\\
        \midrule

        \rowcolor{grdcolor}
        \multicolumn{12}{l}{\textbf{Visual Grounding (4 tasks)}} \\
        MSCOCO & 33.8 & 34.5 & 46.4 & 28.9 & 46.6 & 80.6 & 66.3 & 79.1& 82.4 & 71.2 & 78.1 \\
        \rowcolor{oodcolor}
        RefCOCO & 56.9 & 54.2 & 70.8 & 47.4 & 67.8 &88.7 & 80.8 & 87.4 & 93.3 & 85.4 & 91.1 \\
        \rowcolor{oodcolor}
        RefCOCO-matching & 61.3 & 68.3 & 50.8 & 59.5 & 62.9 & 84.0 & 74.4 & 83.1 & 89.1 & 84.6  & 91.1\\
        \rowcolor{oodcolor}
        Visual7W-pointing & 55.1 & 56.3 & 70.1 & 52.0 & 71.3 & 90.9& 70.0 & 79.9 & 91.8 & 79.2 & 85.8 \\
        \textit{All Visual Grounding} & 51.8 & 53.3 & 59.5 & 47.0  & 65.3 & 86.1 & 72.9 & 82.2 & 89.2 & 80.3 &  86.5\\
        \midrule

        \rowcolor{finalcolor}
        \multicolumn{12}{l}{\textbf{Final Score (36 tasks)}} \\
        All & 37.8 & 39.7 & 34.8 & 25.2 & 44.7 & 62.9& 59.7 & 65.5 & 64.7 & 63.9 & 68.6 \\
        All IND & 37.1 & 39.3 & 32.3 & 25.3 & 47.1 & 67.5 & 65.6 & 71.9 & 70.4 & 69.5 & 74.0 \\
        All OOD & 38.7 & 40.2 & 38.0 & 25.1 & 41.7 &57.1 & 52.3 & 57.5 & 57.5 & 56.9 & 61.8 \\
        \bottomrule
    \end{tabular}
\end{table*}

\end{document}